\documentclass[journal]{IEEEtran}
\usepackage{times}  % DO NOT CHANGE THIS
\usepackage{helvet} % DO NOT CHANGE THIS
\usepackage{courier}  % DO NOT CHANGE THIS
\usepackage[hyphens]{url}  % DO NOT CHANGE THIS
\usepackage{graphicx} % DO NOT CHANGE THIS
\usepackage{caption} % DO NOT CHANGE THIS AND DO NOT ADD ANY OPTIONS TO IT
\usepackage{latexsym}
\usepackage{algorithm}
\usepackage{algorithmic}   
\usepackage{amsmath}
\usepackage{amsfonts}
\usepackage{url}
\usepackage[switch]{lineno}
\usepackage{booktabs}
\usepackage{threeparttable}  
\usepackage{graphicx}
\usepackage{subfigure}
\usepackage{multirow}
\usepackage{amssymb}
\usepackage{placeins}
\usepackage{color}
\usepackage{verbatim}
\usepackage{bm}
\begin{document}

\title{Dif-Fusion: Towards High Color Fidelity in Infrared and Visible Image Fusion with Diffusion Models}
\author{Jun~Yue,
	Leyuan~Fang,~\IEEEmembership{Senior~Member,~IEEE,}
	 Shaobo~Xia,
      Yue~Deng,~\IEEEmembership{Senior~Member,~IEEE,}
  and~Jiayi~Ma,~\IEEEmembership{Senior~Member,~IEEE}
	\thanks{This work has been submitted to the IEEE for possible publication. Copyright may be transferred without notice, after which this version may no longer be accessible.
    \par This work was supported in part by the National Natural Science Foundation of China under Grant U22B2014, Grant 62101072 and Grant 42201481, in part by the Science and Technology Plan Project Fund of Hunan Province under Grant 2022RSC3064.  \emph{(Jun Yue and Leyuan Fang contributed equally to this work.) }}
	\thanks{Jun Yue is with the School of Automation, Central South University, Changsha 410083, China, and also with the Department of Geomatics Engineering, Changsha University of Science and Technology, Changsha 410114, China (e-mail: jyue@pku.edu.cn).}% <-this % stops a space
	\thanks{Leyuan Fang is with the College of Electrical and Information Engineering, Hunan University, Changsha 410082, China, and also with the Peng Cheng Laboratory, Shenzhen 518000, China (e-mail: fangleyuan@gmail.com).}% 
 	\thanks{Shaobo Xia is with the Department of Geomatics Engineering, Changsha University of Science and Technology, Changsha 410114, China (e-mail: shaobo.xia@csust.edu.cn).}% <-this % stops a space
        \thanks{Yue Deng is with the School of Astronautics, Beihang University, Beijing 100083, China (e-mail: yuedeng.thu@gmail.com).}
        \thanks{Jiayi Ma is with the Electronic Information School, Wuhan University, Wuhan 430072, China (e-mail: jyma2010@gmail.com).}
	\thanks{}}

\markboth{}%
{Shell \MakeLowercase{\textit{et al.}}: Bare Demo of IEEEtran.cls for IEEE Journals}

\maketitle

\begin{abstract}
Color plays an important role in human visual perception, reflecting the spectrum of objects. However, the existing infrared and visible image fusion methods rarely explore how to handle multi-spectral/channel data directly and achieve high color fidelity. This paper addresses the above issue by proposing a novel method with diffusion models, termed as Dif-Fusion, to generate the distribution of the multi-channel input data, which increases the ability of multi-source information aggregation and the fidelity of colors. In specific, instead of converting multi-channel images into single-channel data in existing fusion methods, we create the multi-channel data distribution with a denoising network in a latent space with forward and reverse diffusion process. Then, we use the denoising network to extract the multi-channel diffusion features with both visible and infrared information. Finally, we feed the multi-channel diffusion features to the multi-channel fusion module to directly generate the three-channel fused image. To retain the texture and intensity information, we propose multi-channel gradient loss and intensity loss. Along with the current evaluation metrics for measuring texture and intensity fidelity, we introduce a new evaluation metric to quantify color fidelity. Extensive experiments indicate that our method is more effective than other state-of-the-art image fusion methods, especially in color fidelity. The source code will be publicly available at https://github.com/GeoVectorMatrix/Dif-Fusion.
\end{abstract}


\IEEEpeerreviewmaketitle

\section{INTRODUCTION}

\IEEEPARstart{D}{ue} to the theoretical and technical limitations of the optical imaging hardware equipment, the image acquired by a single sensor or a single shooting setting can only obtain part of the image information \cite{ZHANG2021323,9151265}. Therefore, the fusion of images from different sensors or different shooting settings helps to enrich the image information. Among various image fusion tasks, infrared and visible image fusion is one of the most widely used \cite{ma2019infrared,9416507}. The infrared sensor can capture the thermal radiation from the object, but it is vulnerable to noise and difficult to capture the texture information. On the contrary, visible images usually contain rich structure and texture information, but are vulnerable to illumination and occlusion. The complementary between them makes it possible to generate fusion images containing both thermal objects and texture details. Infrared and visible image fusion has been widely used in military, object detection and tracking \cite{li2018cross}, person re-identification \cite{lu2020cross}, semantic segmentation \cite{9531449} and many other fields.

\begin{figure}[t]
	\centering 
{\tiny }	\includegraphics[width=0.49\textwidth]{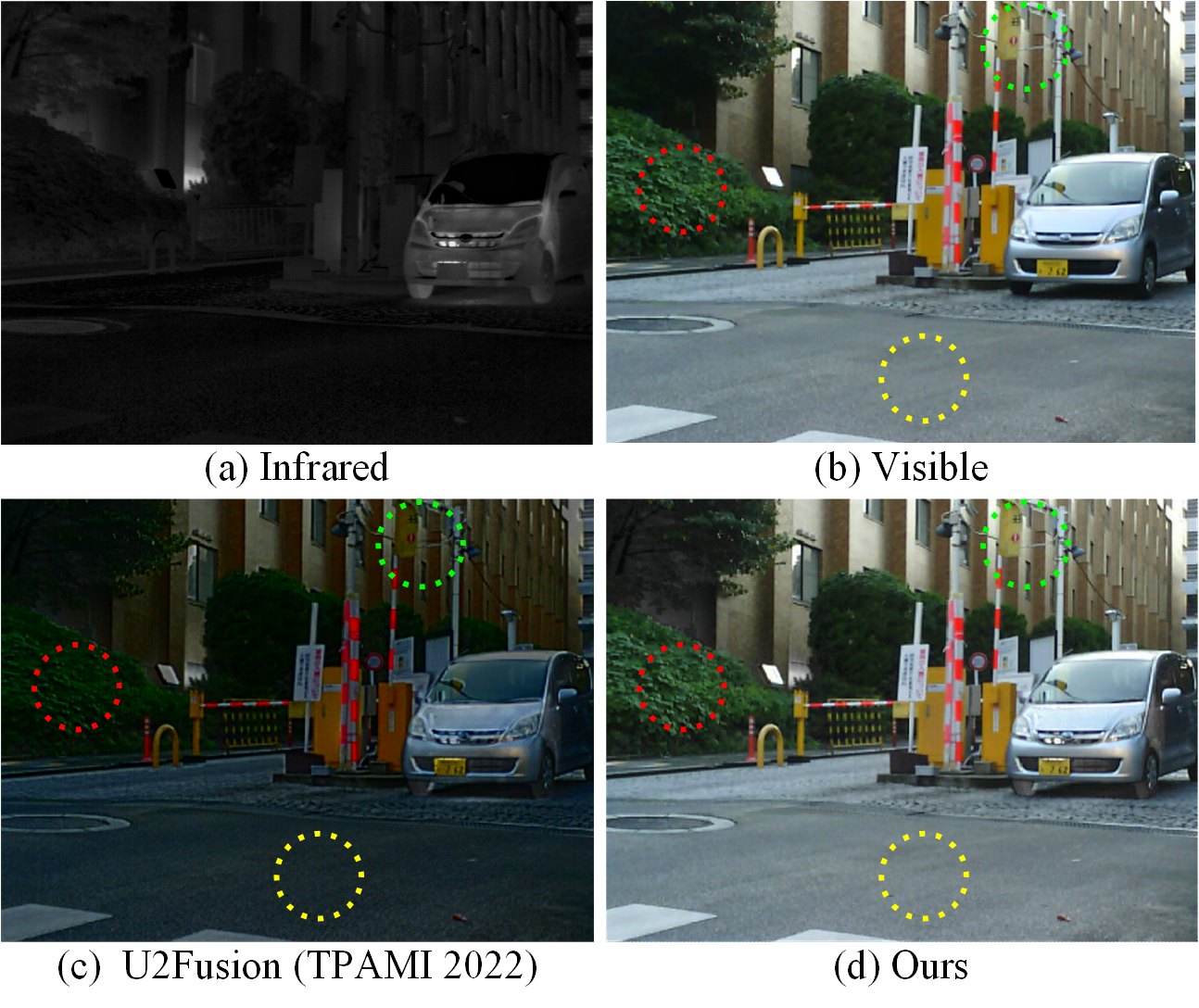} 
	\caption{Illustration of color fidelity. From
(a) to (d): infrared image, visible image, fused images of U2Fusion \cite{9151265} and our proposed Dif-Fusion. The dotted circles in red, yellow, and green show the color differences between the visible and the fused images of the wall, the pavement, and the vegetation, respectively. Compared with existing method, Dif-Fusion achieves higher color fidelity.}
	\label{Weakness} 
\end{figure}

In order to achieve effective fusion of infrared and visible images, many image fusion technologies have been proposed in the past decades \cite{ma2019infrared}, including traditional methods \cite{zhou2016perceptual} and deep learning-based methods \cite{ma2019fusiongan}. Traditional infrared and visible image fusion algorithms can be generally divided into the following categories, including sparse representation-based methods \cite{li2013remote, liu2016image}, multi-scale transformation-based methods \cite{liu2015general,li2013image}, subspace-based methods \cite{kong2014adaptive}, saliency detection-based methods  \cite{bavirisetti2016two}, and hybrid methods \cite {ma2017infrared}. Although the above algorithms can meet the needs of specific scenes in most cases, there are still some problems: 1) The existing traditional methods usually use the same method to express the image features, and rarely consider the distinctive characteristics of infrared and visible images; 2) The activity level measurement and fusion rules need to be set manually, which cannot meet the needs of complex scenarios \cite{li2017pixel}. 

In recent years, with the rapid development of deep learning technology, researchers have explored fusion algorithms based on deep neural networks. Generally, the current mainstream deep fusion methods can be divided into three categories: methods based on autoencoder (AE) \cite{8580578}, methods based on convolutional neural network (CNN) \cite{9416507,TANG202279} and methods based on generative adversarial network (GAN) \cite{ma2019fusiongan,ma2020ganmcc}. As an image generation task, the existing infrared and visible image fusion methods lack in-depth exploration of the generation model. The existing methods based on generation model are mainly based on GAN, including FusionGAN \cite{ma2019fusiongan} and GANMcC \cite{ma2020ganmcc}. However, the distribution of infrared and visible images cannot be built due to the additional constraints that these methods impose on the generator.

Although the existing fusion methods based on deep learning can achieve relatively satisfactory performance, there are still some issues that need to be taken into consideration. First, the existing methods mainly focus on preserving the thermal targets in the infrared image and the background texture structure in the visible image, and less on how to preserve the color information in the visible image \cite{9416507}. However, color reflects the spectrum of objects which is of vital importance in digital images. The human visual system is highly sensitive to color (spectrum) and there are many studies on the significance of color in understanding visual scenes \cite{kingdom2003color,castelhano2008influence}. Extensive theoretical and empirical studies on color clearly indicates that color has a significant influence on people's cognition, affect, and behavior \cite{elliot2014color,mehta2009blue,hill2005red}. As seen in Fig.~\ref{Weakness} (c), the current method (U2Fusion \cite{9151265}) does not effectively utilize multi-spectral information and performs poorly in maintaining the color information in visible images, which will affect human perception negatively. In addition to gradient fidelity and intensity fidelity, we believe that it is also necessary to maintain the color of visible image in the fusion task, which helps to retain information that is critical to human perception \cite{elliot2007color}. 

On the other hand, how to extract multi-channel complementary information within the input data is not well studied. The existing methods usually convert the visible images stored in three channels (i.e., RGB channels) from RGB space to YCbCr space, and use the Y channel for fusion \cite{zhang2021sdnet,9151265}. After the single-channel fused image is generated, it needs to be converted to a three-channel image through post-processing \cite{Liu_2022_CVPR,9623476}. Since not all channels are presented in the input data, it is hard to construct the multi-channel distribution and extract multi-channel complementary information, resulting in color distortion.

To address the above challenges, a novel infrared and visible image fusion method based on diffusion models, namely, Dif-Fusion, is proposed. First, we directly feeds multi-channel data composed of three-channel visible image and one-channel infrared image, and constructs multi-channel distribution in the latent space through diffusion process. The diffusion process is a Markov process, which is divided into a forward process and a reverse process \cite{NEURIPS2020_4c5bcfec}. In the forward process, Gaussian noise is incrementally added to the multi-channel input data, and in the reverse process, the noise added in the forward process is eliminated with multiple timesteps. The multi-channel distribution is constructed by training the denoising network in the reverse process to estimate the noise added in the forward process. Second, we extract the multi-channel diffusion features from the denoising network, which includes both infrared and visible features. Third, the multi-channel diffusion features are fed into the multi-channel fusion module to directly generate three-channel fused images. Furthermore, We propose multi-channel gradient loss $\mathcal{L}_{MCG}$ and multi-channel intensity loss $\mathcal{L}_{MCI}$ to preserve the texture and gradient information of three-channel fused images. 

The existing methods mostly concentrate on fusing texture/gradient in the visible images and intensity in the infrared images, without paying attention to the preservation of color information and the extraction of multi-channel complementary information. The proposed method establishes the distribution of multi-channel input data based on diffusion models and extracts multi-channel complementary information to achieve high color fidelity. As shown in Fig.~\ref{Weakness} (d), our fused image has high color fidelity and is more suitable for human visual perception. In terms of fusion result evaluation, in addition to existing indicators used to quantify intensity and gradient fidelity, we introduce an indicator to quantify color fidelity. With the proposed Dif-Fusion, the infrared and visible images can be simply fed into the model without color space transformation. To sum up, the main contributions of this work are threefold. 
\begin{itemize}
\item We propose an infrared and visible image fusion framework based on diffusion models that can generate chromatic fused image directly and achieve color, gradient and intensity fidelity simultaneously.
\item We formulate the construction of multi-channel distribution as a diffusion process, which is the first study to apply the diffusion models to infrared and visible image fusion to the best of our knowledge.
\item In order to measure the color fidelity of fused images, a new evaluation metric is introduced to quantify the color fidelity. Extensive experiments show the proposed method outperforms the existing state-of-the-art methods.
\end{itemize}

The rest of this paper is organized as follows. In Section II, we briefly introduce the related work of image fusion and diffusion models. In Section III, the proposed method is described in detail. In Section IV, the experimental settings and results are shown and discussed. In Section V, the conclusions of this article are summarized.

\section{Related Work}

In this section, we introduce background materials and related work that are highly relevant to the method proposed in this paper, including traditional infrared and visible image fusion methods, deep learning based fusion methods, and diffusion models.

\subsection{Infrared and Visible Image Fusion}
In the past decades, researchers have proposed many infrared and visible image fusion techniques, including traditional methods and deep learning-based methods \cite{li2017pixel,ZHANG2021323}. Traditional infrared and visible image fusion algorithms can be generally divided into five categories, i.e., sparse representation, multi-scale transformation, subspace representation, saliency detection, and hybrid methods \cite{ma2019infrared}.

The main idea of sparse representation theory is that an image signal can be represented as a linear combination of the least possible atoms or transformation primitives in an overly complete dictionary~\cite{hamida20183}. Over-completeness indicates that the number of atoms in the dictionary is greater than the dimension of the signal \cite{li2013remote, liu2016image}. In image fusion, sparse representation usually learns a complete dictionary from a group of training images, which captures the inherent data-driven image representation. The over-complete dictionary contains abundant base atoms, allowing for more meaningful and stable source image representation~\cite{liu2015general}. Multi-scale transformation can decompose the original image into subimages of different scales \cite{ma2019infrared}. The multi-scale transformation is similar to the human visual process, which can make the fused image have a good visual effect \cite{liu2015general,zhu2017fusion}.

The method based on subspace representation aims to project high dimensional features into low dimensional subspace \cite{mitianoudis2013region}. Projection into low dimensional subspace can help capture the inherent structure of the original input image \cite{fu2016infrared}. In addition, data processing in a low dimensional subspace can save time and memory compared to that in a high-dimensional space. Common subspace representation-based methods include principal component analysis (PCA) \cite{li2016improved,8009719}, independent component analysis (ICA) \cite{cvejic2007region,mitianoudis2007pixel} and non-negative matrix factorization (NMF) \cite{mou2013image,kong2014adaptive}. 

The saliency detection model simulates human behavior and captures the most prominent regions/objects from images or scenes \cite{zhao2019pyramid}. It has many important applications in computer vision and pattern recognition tasks \cite{ullah2020brief,wang2020large}. In recent years, infrared and visible image fusion methods based on saliency detection can be mainly divided into two categories, namely, weight calculation \cite{bavirisetti2016two,ma2017infrared,cui2015detail} and salient target extraction \cite{han2013fast,zhang2015fusion,liu2017infrared}. The above fusion methods all have advantages and disadvantages. Researchers explore hybrid methods to make full use of the advantages of various methods and improve image fusion performance. Common hybrid methods include hybrid multi-scale transformation and sparse representation \cite{yin2015sparse,liu2015general}, hybrid multi-scale transformation and saliency detection \cite{zhao2013fusion,bavirisetti2016two}, etc.

Due to the excellent feature learning ability and nonlinear fitting ability of neural networks, researchers have explored data-driven infrared and visible image fusion methods based on deep learning \cite{ZHANG2021323,9970457,9812535}. These methods mainly include AE-based methods \cite{8580578,9127964,9187663}, CNN-based methods \cite{Zhang_Xu_Xiao_Guo_Ma_2020,zhang2021sdnet,9416507}, and GAN-based methods \cite{ma2019fusiongan,9031751,ma2020ganmcc}.

Researchers have proposed many fusion methods based on AE \cite{ma2019infrared}. Most of them use the encoder structure to extract features from the source image, and use decoder structure to complete image reconstruction. DenseFuse \cite{8580578} is a typical AE-based method. The encoding network of this method is combined with convolution layers, fusion layer, and dense blocks, where the output of each layer is connected with every other layer. The fused image is then reconstructed by a decoder. In order to improve the feature extraction ability of the encoder, researchers proposed a method named NestFuse \cite{9127964}. From a multi-scale perspective, this method can preserve a large amount of information from the input data based on nest connections. SEDRFuse \cite{9187663} is a symmetric encoder-decoder with residual networks. In the fusion phase, the trained extractor is used to extract intermediate and compensated features, and then two attention maps derived from the intermediate features are multiplied by the intermediate features for fusion \cite{9187663}. 

\begin{figure*}[t]
	\centering 
{\tiny }	\includegraphics[width=\textwidth]{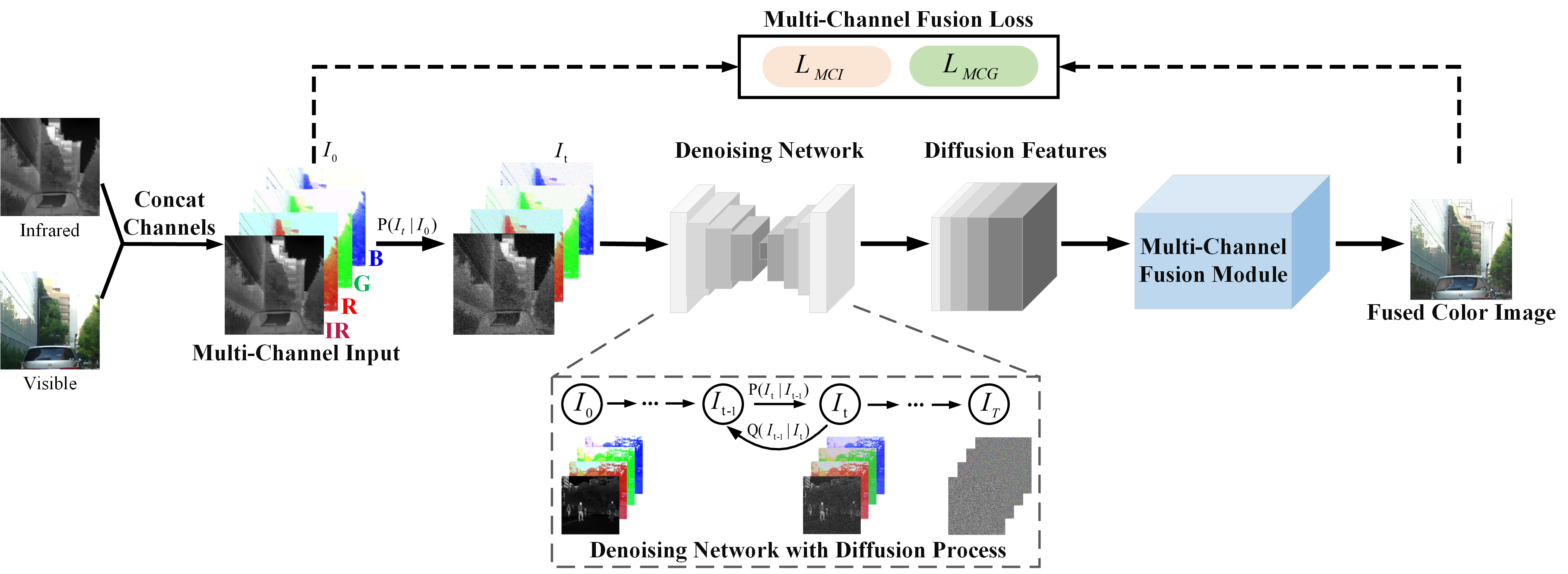} 
	\caption{The overall framework of Dif-Fusion. $\bm{I_{0}}$ and $\bm{I_{t}}$ denote the multi-channel input and the multi-channel data in the forward diffusion process with $t$ timesteps. $P(\cdot|\cdot)$ and $Q(\cdot|\cdot)$ stand for the forward diffusion possess and reverse diffusion process. $\mathcal{L}_{MCI}$ and $\mathcal{L}_{MCG}$ represent multi-channel gradient loss and multi-channel intensity loss.} 
	\label{fig:idea} 
\end{figure*}

For CNN-based fusion methods, a typical method is PMGI \cite{Zhang_Xu_Xiao_Guo_Ma_2020}. This method is a fast unified image fusion network based on gradient and intensity proportion preserving. At the same time, the method introduces a pathwise transfer block to exchange information between different paths, which can pre-fuse gradient and intensity information to enhance the information to be fused. In order to adaptively determine the proportion of gradient information preserved and retain more complete texture structure, SDNet is proposed \cite{zhang2021sdnet}. For gradient fidelity, this method determines the optimization objective of gradient distribution according to the texture richness, and guides the fusion image to contain more texture details through an adaptive decision block. To provide spatial guidance for the integration of multi-source information, STDFusionNet employs a salient target mask to help with the fusion task \cite{9416507}. To combine fusion tasks with a high-level visual task, a fusion method assisted by a high-level semantic task is proposed \cite{tang2022image}. In addition, there are a few methods to investigate how lighting condition affects image fusion \cite{TANG202279,tang2023divfusion}.

Since GAN has the ability to estimate the probability distribution in an unsupervised manner, some image fusion techniques based on GAN are proposed \cite{ZHANG2021323,ma2019infrared}. Among them, FusionGAN \cite{ma2019fusiongan} creates an adversarial game between the generator and the discriminator. The generator aims to generate the fused image, while the discriminator attempts to force the fused image to contain more details of the visible image \cite{ma2019fusiongan}. In order to address the problem that the discriminator is only used to distinguish visible images, a dual-discriminator conditional generative adversarial network (DDcGAN) is proposed, which uses two discriminators to identify the structural differences between the fused image and the source images \cite{9031751}. To help the generator focus on the foreground object information of the infrared image and the background details of the visible image, researchers exploit multi-scale attention mechanisms to fuse infrared and visible images for both generator and discriminator (AttentionFGAN) \cite{9103116}. To balance the information between infrared and visible images, a fusion method named generative adversarial network with multiclassification constraints (GANMcC) is proposed \cite{ma2020ganmcc}. However, the fusion methods mentioned above based on the generation model, whose generators add gradient fidelity and intensity fidelity constraints during training, cannot realize the distribution construction of infrared and visible images in the latent space. At the same time, the existing methods usually convert the three-channel visible image into a single channel image, making it challenging to fully utilize the multi-spectral information and achieve high color fidelity.

\subsection{diffusion models}

diffusion models have become a powerful family of deep generation models \cite{yang2022diffusion,nichol2021improved}, with record breaking performance in many areas \cite{song2020denoising}, including image generation \cite{sohl2015deep,NEURIPS2020_4c5bcfec,song2020score,dhariwal2021diffusion}, image inpainting \cite{lugmayr2022repaint}, image super-resolution \cite{saharia2022image,daniels2021score,chung2022come}, and image-to-image translation \cite{saharia2022palette,zhao2022egsde,wolleb2022swiss}. In addition, the feature representations learned from the diffusion models are also found to be very useful in discriminative tasks, including image classification \cite{zimmermann2021score}, image segmentation \cite{baranchuk2021label,amit2021segdiff} and object detection \cite{chen2022diffusiondet}. A diffusion models is a deep generative model with two processes, namely the forward process and the reverse process \cite{croitoru2022diffusion}. In the forward process, the input data is gradually disturbed in several timesteps by adding Gaussian noise. In the reverse process, the task of the model is to recover the original input data through multiple reverse timesteps by reducing the difference between the added noise and the predicted noise \cite{baranchuk2021label}.

diffusion models are widely used to generate samples because of the high quality and variety of samples generated by the models \cite{croitoru2022diffusion}. With its continuous development in various fields, diffusion models break the long-term dominant position of the generation adversarial network in the image generation field \cite{yang2022diffusion}. The fusion of infrared and visible images can also be regarded as an image generation task. This paper explores an effective way to achieve state-of-the-art fusion results with diffusion models.

\section{Method}
In this section, we describe the diffusion-based image fusion framework for multimodal data in detail. The main idea of the proposed method is illustrated in Fig.~\ref{fig:idea}. The visible image and infrared image pairs are concatenated along the channel dimension to make a multi-channel input for the diffusion models. In the forward process, Gaussian noise is gradually added to the multi-channel data until the data is close to pure noise (e.g., $P(\bm{I_{t}}|\bm{I_{t-1}})$), as shown in Fig.~\ref{fig:idea}. Then, the reverse process tries to predict and remove the added noise with the help of a denoising network (e.g., $Q(\bm{I_{t-1}}|\bm{I_{t}})$). After that, diffusion features can be extracted from the diffusion models and fed to the proposed multi-channel fusion module, as shown in Fig.~\ref{fig:idea}. The chromatic fusion images will be produced by the proposed framework directly under the guidance of the proposed multi-channel losses.

\par In the following subsections, we first go through how diffusion models learn the multi-channel distribution and generate new image pairs. Next, a multi-source information aggregation method based on diffusion models is presented in detail. Finally, we introduce multi-channel intensity loss and multi-channel gradient loss to guide the fusion network's training process.

\subsection{Joint Diffusion with Infrared and Visible Images}
Given a pair of registered infrared image $\bm{I_{ir}}\in \mathbb{R}^{H×W×1}$ and visible image $\bm{I_{vis}}\in \mathbb{R}^{H×W×3}$, where $H$ and $W$ represent the height and width, respectively. In order to learn the joint latent structure of multi-channel data, the 1-channel infrared image and 3-channel visible image are concatenated to form a 4-channel image, which is represented by a $\bm{I}\in \mathbb{R}^{H×W×4}$. We adopt the diffusion process proposed in Denoising Diffusion Probabilistic Model (DDPM) \cite{NEURIPS2020_4c5bcfec} to construct the distribution of multi-channel data. The forward diffusion process of the multi-channel image is to gradually add noises with $T$ timesteps. In the reverse process, the noise is gradually eliminated through $T$ timesteps. The goal of training the diffusion models with the forward and reverse process is to learn the joint latent structure of the infrared and visible images by modeling the diffusion of the 4-channel images in the latent space \cite{NEURIPS2020_92c3b916,Gu_2022_CVPR}.

\subsubsection{Forward Diffusion Process}
The forward diffusion process inspired by non-equilibrium thermodynamics \cite{sohl2015deep} can be viewed as a Markov chain that gradually adds Gaussian noise to the data with $T$ timesteps \cite{NEURIPS2020_4c5bcfec}. At timestep $t$, the noisy multi-channel image $\bm{I_{t}}$ can be represented as follows:
\begin{eqnarray}
P(\bm{I_{t}}|\bm{I_{t-1}})=\mathcal{N}(\bm{I_{t}};\sqrt{\alpha_{t}}\bm{I_{t-1}},(1-\alpha_{t})\bm{Z})
	\label{xt}
\end{eqnarray}
where $\bm{Z}$ denotes the standard normal distribution. $\bm{I_{t}}$ and $\bm{I_{t-1}}$ represents the noisy 4-channel images generated by adding Gaussian noises for $t$ and $t-1$ times, respectively. $\bm{\gamma} \in \mathbb{R}^{H×W×4}$ is the Gaussian noise. $\alpha_{t}$ is the variance schedule that controls the variance of the Gaussian noise added in timestep $t$. More specifically, for the first timestep, the noisy 4-channel image $\bm{I_{1}}$ can be formulated as:
\begin{eqnarray}
	\bm{I_{1}}=\sqrt{\alpha_{1}}\bm{I_{0}}+\sqrt{1-\alpha_{1}}\bm{\gamma}
	\label{x1}
\end{eqnarray}
where $\bm{I_{0}}$=$\bm{I}$. Given the original input $\bm{I_{0}}\in \mathbb{R}^{H×W×4}$, the expression of $\bm{I_{t}}$ can be deduced by Eq.~\eqref{xt} and Eq.~\eqref{x1}:
\begin{eqnarray}
P(\bm{I_{t}}|\bm{I_{0}})=\mathcal{N}(\bm{I_{t}};\sqrt{\bar{\alpha}_{t}}\bm{I_{0}},(1-\bar{\alpha}_{t})\bm{Z})
	\label{x0xt}
\end{eqnarray}
where $\bar{\alpha}_{t}=\prod \limits_{i=1}^t\alpha_{i}$. In the forward diffusion process, given the timestep $t$, variance schedule $\alpha_{1},...,\alpha_{t}$, and the sampled noise, the noisy multi-channel sample of timestep $t$ can be directly calculated by Eq.~\eqref{x0xt}.

\begin{figure}[t]
	\centering 
	\includegraphics[width=0.49\textwidth]{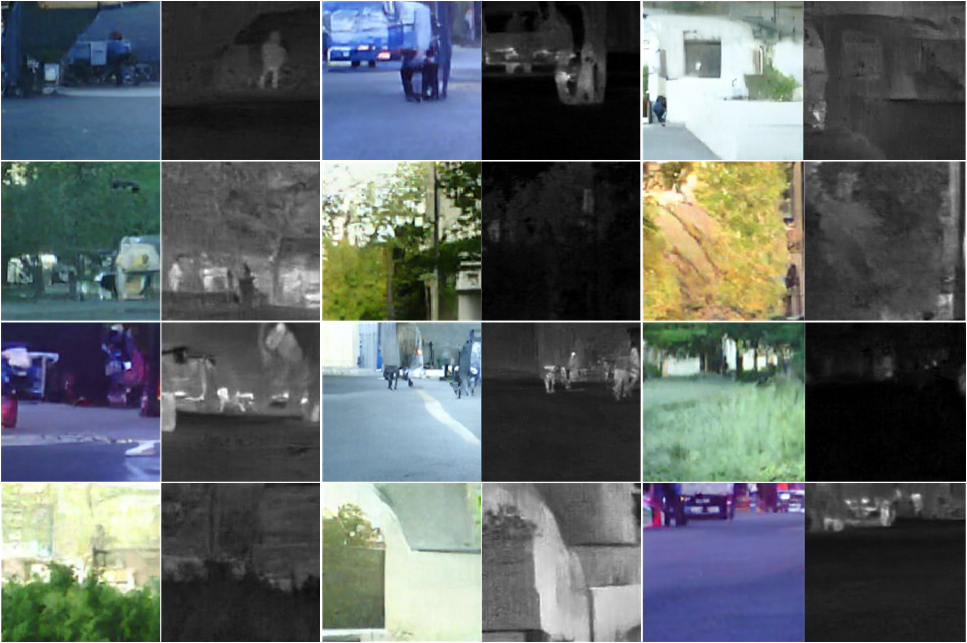} 
	\caption{Visible and infrared image pairs generated from the diffusion models.}
	\label{Fig:generated} 
\end{figure}

\subsubsection{Reverse Diffusion Process}
In the reverse diffusion process, neural networks are used to perform a series of small denoising operations to obtain the original multi-channel image \cite{baranchuk2021label}. In each timestep of the reverse process, the denoising operation is performed on the noisy multi-channel image $\bm{I_{t}}$ to obtain the previous image $\bm{I_{t-1}}$ \cite{gedara2022remote}. The probability distribution of $\bm{I_{t-1}}$ under the condition $\bm{I_{t}}$ can be formulated as \cite{NEURIPS2020_4c5bcfec}:
\begin{eqnarray}
	Q(\bm{I_{t-1}}|\bm{I_{t}})=\mathcal{N}(\bm{I_{t-1}};\mu_{\theta}(\bm{I_{t}},t),\sigma_{t}^{2}\bm{Z})
	\label{xtxt-1}
\end{eqnarray}
where $\sigma_{t}^{2}$ is the variance of the conditional distribution $Q(\bm{I_{t-1}}|\bm{I_{t}})$, which can be formulated as:
\begin{eqnarray}
	\sigma_{t}^{2}=\frac{1-\bar{\alpha}_{t-1}}{1-\bar{\alpha}_{t}}\beta_{t}
	\label{variance}
\end{eqnarray}
where $\beta_{t}=1-\alpha_{t}$. The mean $\mu_{\theta}(\bm{I_{t}},t)$ of the conditional distribution $Q(\bm{I_{t-1}}|\bm{I_{t}})$ can be formulated as:
\begin{eqnarray}
	\mu_{\theta}(\bm{I_{t}},t)=\frac{1}{\sqrt{\alpha_{t}}}(\bm{I_{t}}-\frac{\beta_{t}}{\sqrt{1-\bar{\alpha}_{t}}}\epsilon_{\theta}(\bm{I_{t}},t))
	\label{mean}
\end{eqnarray}
where $\epsilon_{\theta}(\cdot,\cdot)$ is the denoising network. The inputs of $\epsilon_{\theta}(\cdot,\cdot)$ are the timestep $t$ and the noisy multi-channel image $\bm{I_{t}}$.

\subsubsection{Loss Function of Diffusion Process}
First, we sample a pair of registered visible and infrared image pairs $(\bm{I_{ir}},\bm{I_{vis}})$ in the training set to form the multi-channel image $\bm{I}$. Then we sample the noise $\bm{\gamma}$ from the standard normal distribution. Third, we sample the timestep $t\sim U(\{1,...,T\})$ from the uniform distribution. After completing the above sampling, the loss function of the diffusion models can be formulated as:
\begin{eqnarray}
	\mathcal{L}_{diff}=\left \| \bm{\gamma}-\epsilon_{\theta}(\sqrt{\bar{\alpha}_{t}}\bm{I_{0}}+\sqrt{1-\bar{\alpha}_{t}}\bm{\gamma},t)\right \|_{2}
	\label{mean}
\end{eqnarray}

\subsubsection{Structure of the Denoising Network}
In order to predict the noise added in the forward diffusion process, the structure of the denoising network $\epsilon_{\theta}(\cdot,\cdot)$ adopts the U-Net structure used in SR3 \cite{saharia2022image}. The SR3 backbone consists of a contracting path, an expansive path and a diffusion head. The contracting path and the expansive path are composed of 5 convolution layers. The diffusion head consists of a single convolution layer, which is used to generate the predicted noise.
\par 
Fig. \ref{Fig:generated} shows some paired visible and infrared images generated by our trained diffusion models. These image pairs can be seen to visually resemble the real visible and infrared images. The targets that are highlighted in the corresponding infrared images also appear plausible. These results demonstrate that the diffusion models is a powerful tool for constructing the distributions of multi-channel data.

\subsection{Fusion with Multi-channel Diffusion Features}
After training the denoising network, we use the denoising network to extract the multi-channel features. In the image fusion training stage, we use two kinds of losses (i.e., multi-channel gradient loss and multi-channel intensity loss) for training. It is worth noting that by using the multi channel loss, the three-channel fused images can be generate directly without color space transformation.
\subsubsection{Multi-channel Diffusion Features}
\begin{figure}[t]
	\centering 
	\includegraphics[width=0.49\textwidth]{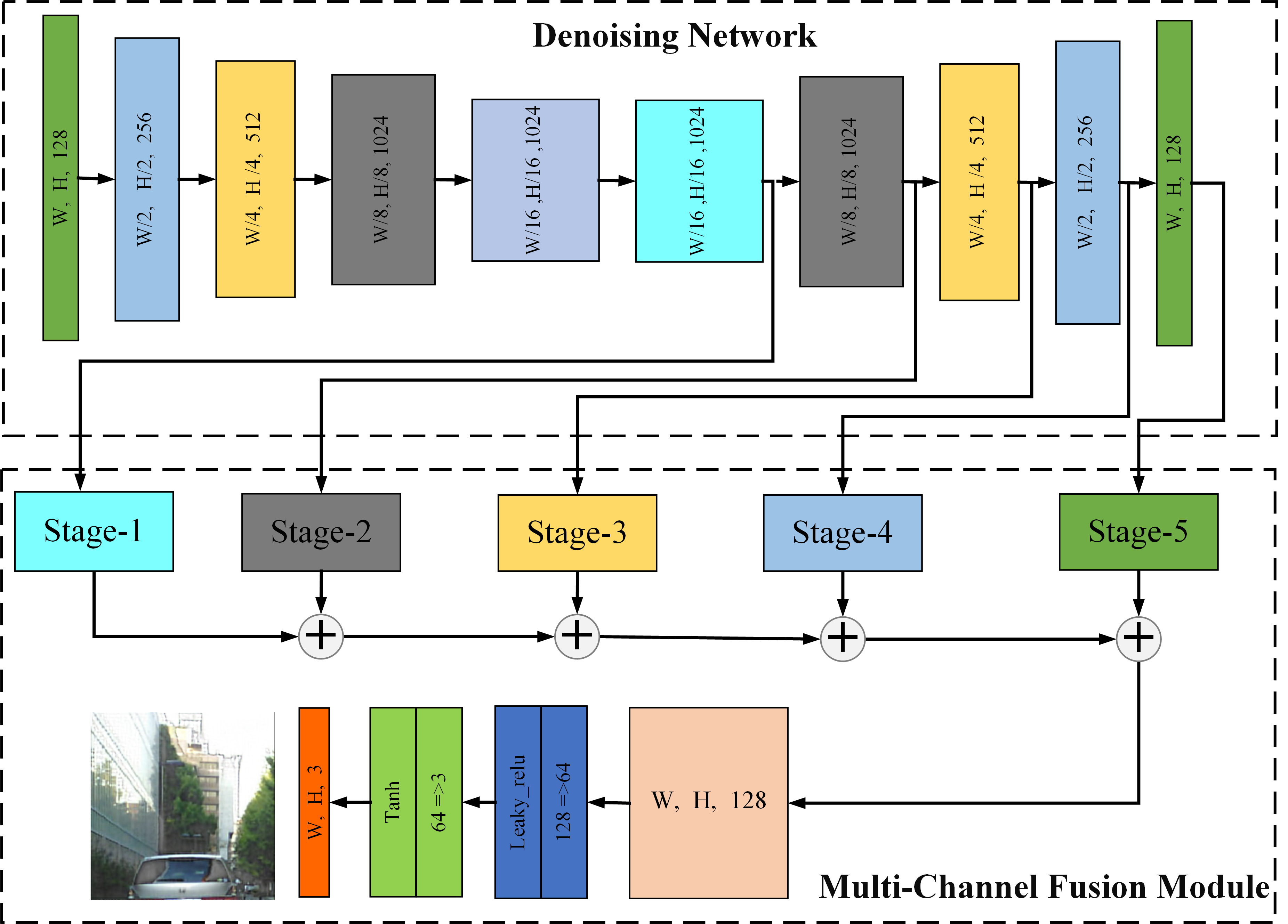} 
	\caption{The structure of the denoising network and the multi-channel fusion module.} 
	\label{fig:overall} 
\end{figure}

For the SR3 backbone, its expansive path contains five convolution layers, and the sizes of its output feature maps are $W/16×H/16, W/8×H/8, W/4×H/4, W/2×H/2, W×H$. We use a multi-channel fusion module to fuse multi-channel diffusion features from 5 stages of the denoising network~\cite{baranchuk2021label,gedara2022remote}. For the five stage features of the five expansive layers, we add them up and feed them into the fusion head to generate the fused image $\bm{I_{f}}\in \mathbb{R}^{H×W×3}$. Specifically, 3 × 3 convolutional layers are applied to map high-dimensional fused features to 3-channel outputs. Leaky ReLU and Tanh are adopted as the activation function. The structure of the denoising network and the multi channel fusion module is shown in Fig.~\ref{fig:overall}. 

\subsubsection{Loss Function of Fusion Process}
Since the visible image has abundant texture information, in order to retain sufficient texture information in the final fused image, we apply a gradient loss for gradient fidelity. However, the existing gradient loss is designed for single-channel fused images \cite{tang2022image}. To directly generate three-channel fused image while keeping the gradient, we extends the existing gradient loss and proposes multi-channel gradient loss $\mathcal{L}_{MCG}$, which can be formulated as:
\begin{eqnarray}
	\mathcal{L}_{MCG}=\frac{1}{HW}\sum_{i=1}^3\left \| \nabla \bm{I_{f}^{i}} - \mbox{max}(\nabla \lvert \bm{I_{ir}} \lvert,\nabla \lvert \bm{I_{vis}^{i}} \lvert ) \right \|_{1}
	\label{spatialloss2}
\end{eqnarray}
where $\nabla$ represents the gradient operator. $\bm{I_{f}^{1}}$, $\bm{I_{f}^{2}}$ and $\bm{I_{f}^{3}}$ represent the there channels (i.e., red, green and blue) of the fused image $\bm{I_{f}}$. $\bm{I_{vis}^{1}}$, $\bm{I_{vis}^{2}}$ and $\bm{I_{vis}^{3}}$ denote the there channels of the input visible image $\bm{I_{vis}}$.
The thermal radiation is usually characterized by pixel intensity \cite{ma2016infrared}. We apply intensity loss to make the fused image have a intensity distribution similar to the infrared image and the visible image. However, similar to gradient loss, the current intensity loss is designed for generating single channel fused images \cite{tang2022image}. We extend the existing intensity loss into multi-channel intensity loss $\mathcal{L}_{MCI}$, which can be formulated as:
\begin{eqnarray}
	\mathcal{L}_{MCI}=\frac{1}{HW}\sum_{i=1}^3\left \| \bm{I_{f}^{i}}- \mbox{max}(\bm{I_{ir}},\bm{I_{vis}^{i}}) \right \|_{1}
	\label{spatialloss2}
\end{eqnarray}

Existing fusion methods usually preserve color information through color space conversion. In order to solve this problem and make full use of diffusion features, this paper directly generates three-channel fused images with multi-channel gradient and intensity losses. The final loss $\mathcal{L}_{f}$ can be formulated as:
\begin{eqnarray}
	\mathcal{L}_{f}=\mathcal{L}_{MCG}+\mathcal{L}_{MCI}
	\label{spatialloss2}
\end{eqnarray}

\section{Experiments}
In this section, we first describe the experiment details, including datasets, evaluation metrics, and the training process. Then, we conduct quantitative and qualitative analysis on three public datasets to evaluate the proposed framework. Also, we compare the performance of our method with six state-of-the-art models in order to demonstrate the benefits of Dif-Fusion. Finally, we reveal the effectiveness and advantages of using diffusion models in multi-channel information fusion based on the ablation study.
\subsection{Experimental Settings}
\subsubsection{Datasets}
We utilize the color and infrared image pairs from the MSRS~\cite{TANG202279}, RoadSence~\cite{9151265}, and M3FD datasets~\cite{liu2022target} to evaluate the proposed framework. We also compare our method with six state-of-the-art algorithms: FusionGAN~\cite{ma2019fusiongan}, SDDGAN~\cite{9623476}, GANMcC~\cite{ma2020ganmcc}, SDNet~\cite{zhang2021sdnet}, U2Fusion~\cite{9151265}, and TarDAL~\cite{liu2022target}. SDNet and U2Fusion are fusion approaches based on CNN architectures, while FusionGAN, SDDGAN, GANMcC and TarDAL are based on generative models and their variants. For the methods that are being compared, fused images are generated using publicly accessible codes and pre-trained models. To produce color results for visual analysis and quantitative evaluation, those single-channel fused results from comparison methods will be converted to color images in post-processing.
\subsubsection{Evaluation Metrics}
Six statistical metrics are used in the quantitative evaluation, five of which are virtual information (MI)~\cite{qu2002information}, visual information fidelity (VIF)~\cite{han2013new}, spatial frequency (SF)~\cite{eskicioglu1995image}, Qabf~\cite{xydeas2000objective}, and standard deviation (SD). MI primarily assesses how well the information from the initial image pairs has been aggregated in the fused image. VIF evaluates the fidelity of the information present in the fused image. The spatial frequency-related information in the combined data is measured by SF. The edge information from the source images is quantified using Qabf. SD primarily evaluates the contrast of composite images.
\par Specifically, we introduced the Delta E~\cite{sharma2005ciede2000}, a color difference calculation index built in CIELAB space that is believed to be more in line with the human perception system~\cite{backhaus2011color}, to quantify the color distortion between the fused image and the original visible image. Delta E is a sort of color distance measurement. As the human eye is more sensitive to some colors than others due to perceptual non-uniformities, the Euclidean distance directly measured in the color space does not match human perception~\cite{sharma2005ciede2000}. The Delta E is recommended as a solution to these problems, along with several corrections for neutral colors, lightness, chroma, hue, and  hue rotation~\cite{backhaus2011color}. 
\par It should be noted that whereas other measures require the original images, SF and SD metrics can be calculated directly on the fused images. A lower Delta E value suggests smaller color distortion and better fusion quality, but the other five metrics work in reverse, with a higher value indicating a better fusion result.

\subsubsection{Training Details}
The proposed model is trained on the MSRS dataset, which includes 1083 training pairs of visible and infrared images. There are 361 pairs of test images in the MSRS dataset. To train the diffusion models, we adopt the training settings in ~\cite{gedara2022remote,saharia2022image}. Specifically, we randomly crop $160 \times 160$ patches from visible and infrared images in the training process. Inspired by the recent work~\cite{baranchuk2021label, gedara2022remote}, we extract diffusion features generated at three time-steps (e.g., 5, 50, 100) to form multi-channel diffusion features. When training the fusion module, the Adam optimizer is utilized to minimize loss, and the learning rate is set to 0.0001. We set the batch size to 24 and the model is trained for 300 epochs. In the test, the visible and infrared images are fed to the networks with the original size. The outputs of our model are color images, which are directly used in qualitative and quantitative analyses. The proposed model is implemented based on PyTorch. All the experiments involved were carried out on a workstation containing the NVIDIA RTX3090 GPU and 3.80 GHz Intel (R) Core (TM) i7-10700K CPU.

\subsection{Fusion Performance Analysis}

\subsubsection{Qualitative Results}
\par  The MSRS dataset consists of both daytime and nighttime scenarios. To demonstrate the advantages of our method in complementary information fusion, texture preservation, and color fidelity, we select two image pairs from each of the two scenarios to show the results from different models. 
Infrared images highlight objects in daytime scenes that have high thermal radiation information, whereas visible images contain rich texture and color information. We hope that the fused color images will be able to emphasize the significant targets in the infrared images while preserving the original visible image's fine-grained texture and color information.

\begin{figure}[t]
	\centering 
	\includegraphics[width=0.49\textwidth]{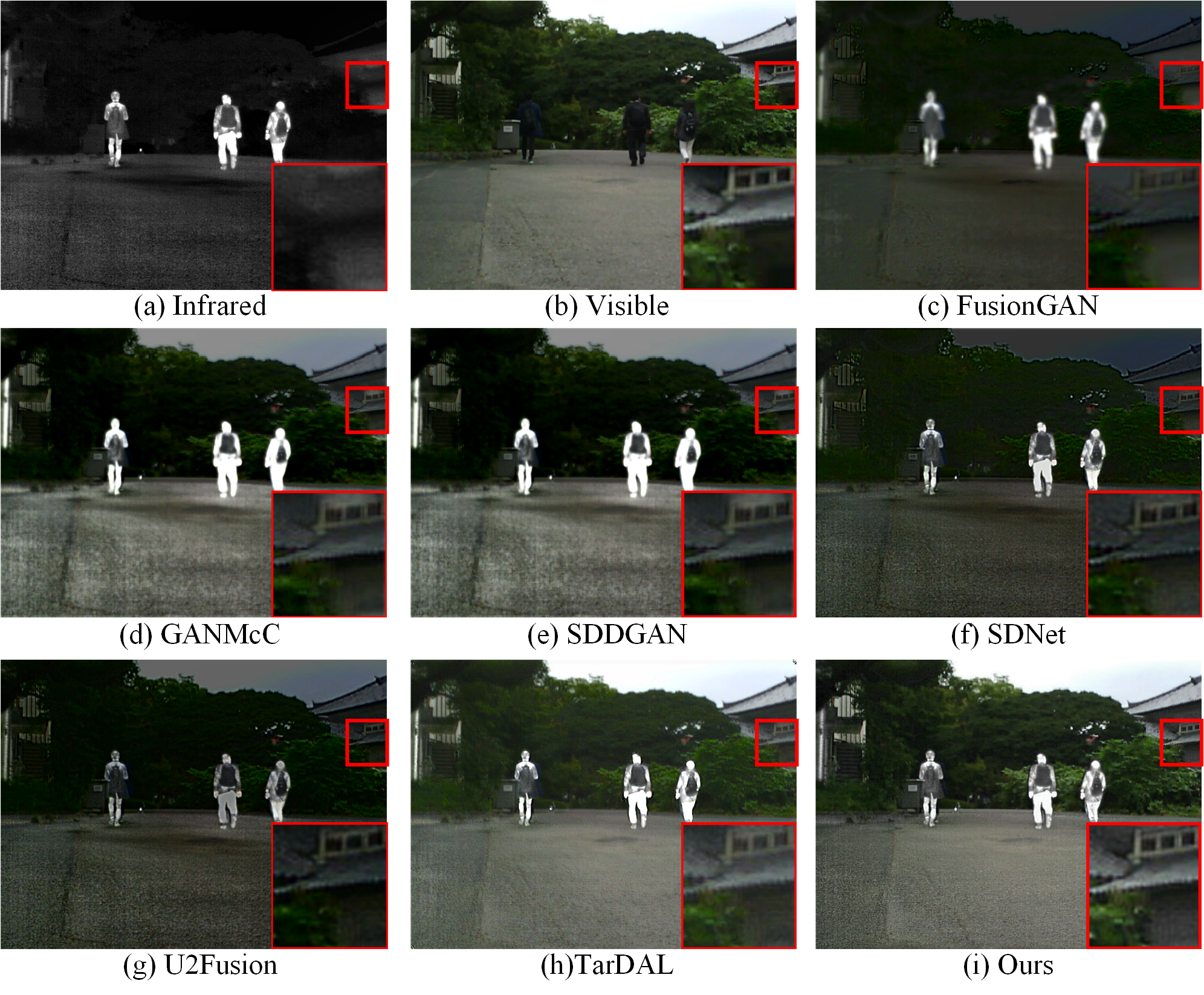} 
	\caption{Qualitative comparison of Dif-Fusion with six state-of-the-art methods on the 00634D image pair from the MSRS dataset.} 
	\label{Fig:MSRS1} 
\end{figure}

\par The infrared image in Fig.~\ref{Fig:MSRS1} highlights three pedestrians, which are persevered in the fused images generated by all methods. However, only the results of our method and TarDAL closely resemble the original visible image. The composited images obtained by other methods (such as FusionGAN, GANMcC, etc.) are visually darker and have large color distortions, such as green trees turn into black in the fused images produced by SDDGAN, U2Fuison, SDNet. The red box in Fig.~\ref{Fig:MSRS1} enlarges the details of the windows under the eaves to demonstrate the benefits of our method in detail maintenance. Only the results from FusionGAN, SDDGAN, TarDAL, and our method represent the fact that the area beneath the eaves in the infrared image has a marginally higher brightness than the surroundings. But only our approach can clearly maintain the window's distinctive contours and arrangements. Additionally, our method makes it easy to discern between the foreground (greenery) and the background (walls beneath windows) in the fused image. 

\par
\begin{figure}[t]
	\centering 
	\includegraphics[width=0.49\textwidth]{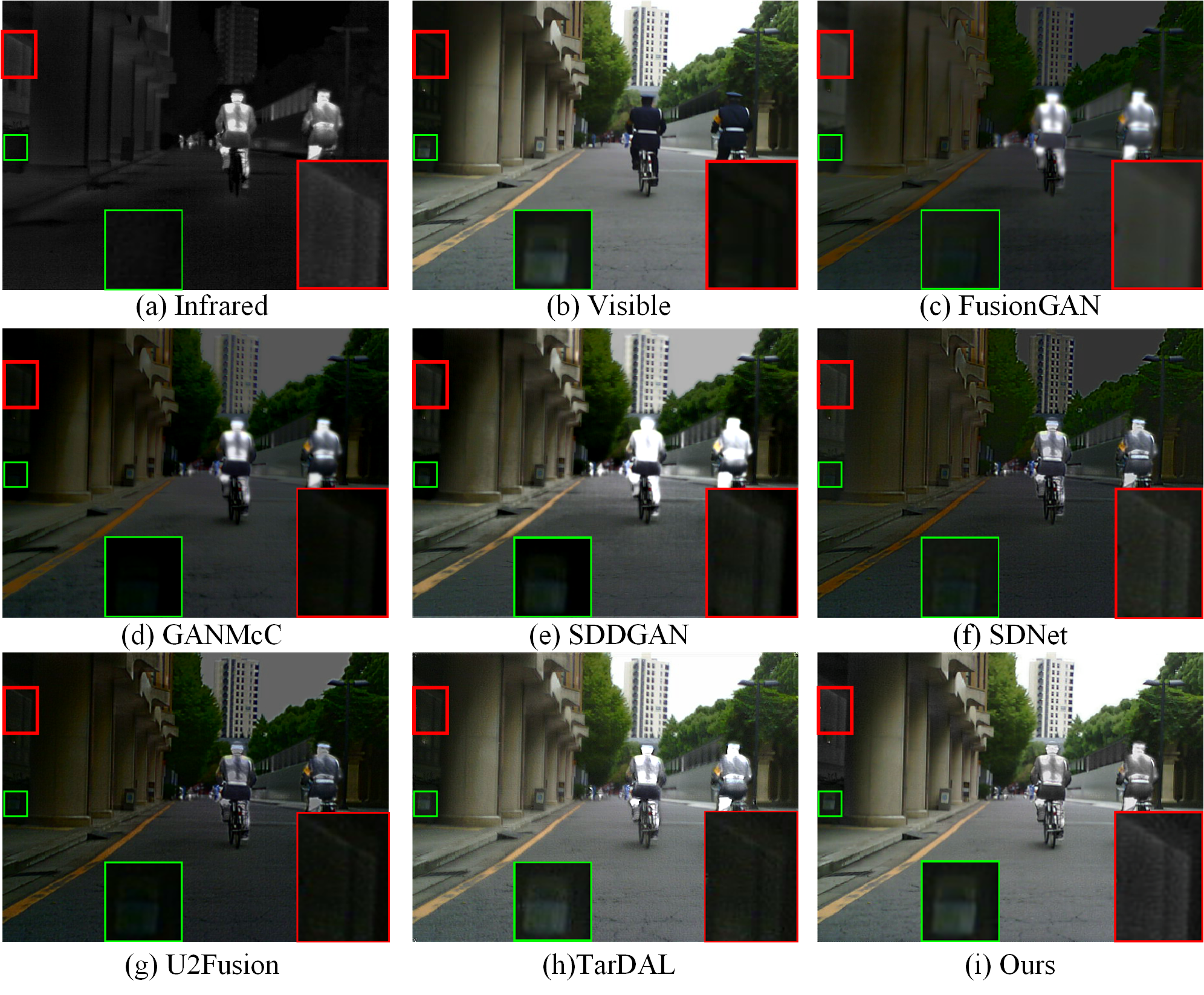} 
	\caption{Qualitative comparison of Dif-Fusion with six state-of-the-art methods on the 00537D image pair from the MSRS dataset.} 
	\label{Fig:MSRS2} 
\end{figure}

 Another pair of daytime images is shown in Fig.~\ref{Fig:MSRS2}. The infrared image primarily shows two bicyclists and some distant pedestrians as highlighted targets. This feature is apparent in the composite images produced by all approaches. Like Fig.~\ref{Fig:MSRS1}, the results from TarDAL and our approach visually resemble the original visible image more closely. The regions in the red and green rectangles have been enlarged. The window structure is only conspicuously visible in the infrared image and not in the visible image. The green region, on the other hand, has a white sign that is only visible in the visible image and not in the infrared image. Some methods (e.g., FusionGAN, GANMcC) struggle to display these features clearly. U2Fusion and SDNet can display the structural information in the red box, while SDDGAN and TarDAL can emphasize the information in the green box. However, only our approach can simultaneously maintain the crucial characteristics in both rectangles. The analysis above reflects the advantages of our method in terms of color preservation, learning complementary information, characterizing details, etc.

\par
\begin{figure}[t]
	\centering 
	\includegraphics[width=0.49\textwidth]{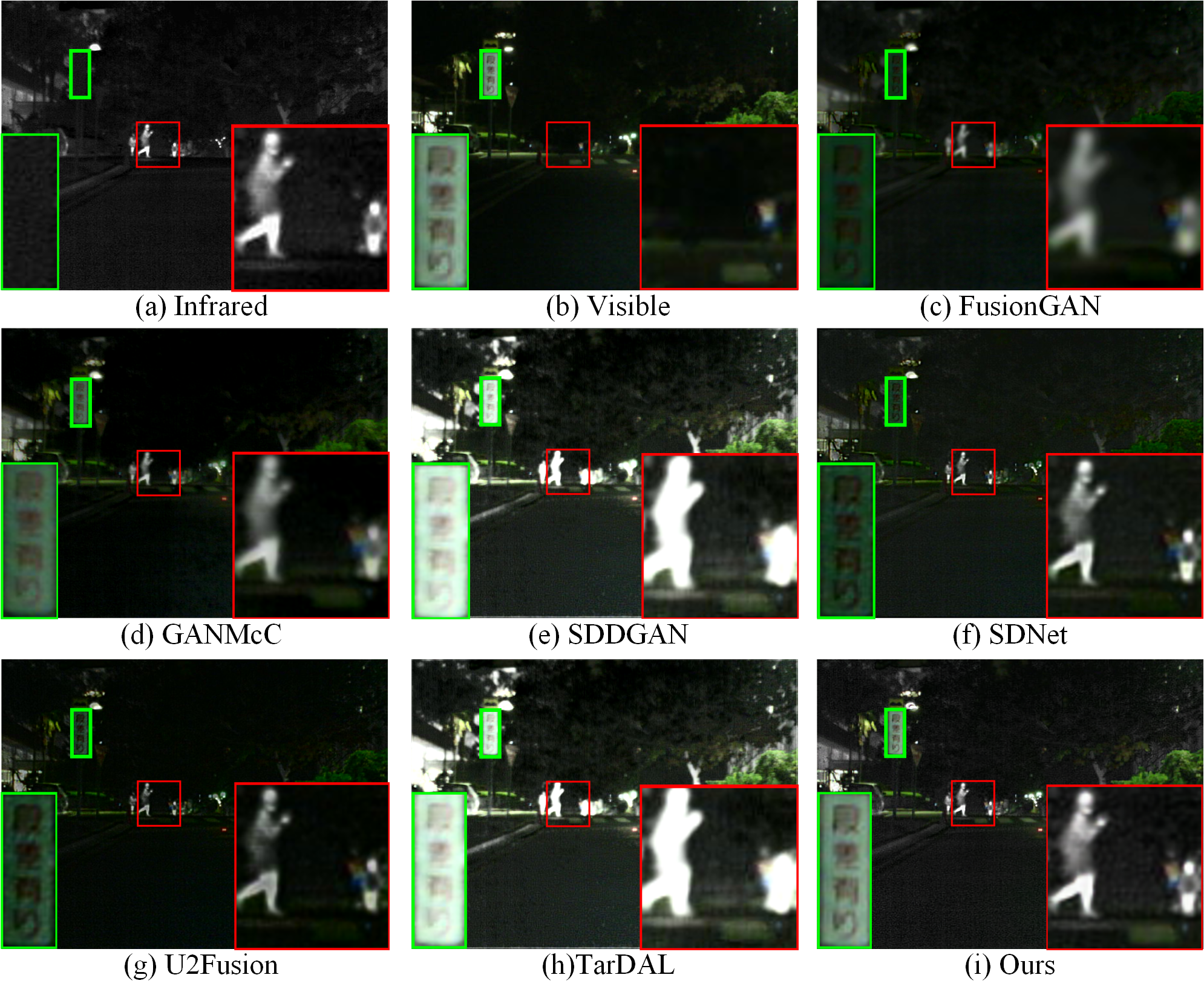} 
	\caption{Qualitative comparison of Dif-Fusion with six state-of-the-art methods on the 00878N image pair from the MSRS dataset.} 
	\label{Fig:MSRS3} 
\end{figure}

 In nighttime scenarios, the infrared image still highlights the target with active thermal radiation, whereas the visible image only has rich color and texture when there is adequate illumination. Therefore, we anticipate that the fused three-channel image will be able to preserve the valuable information in the visible image as well as the infrared image's highlighted targets. We enlarged the red and green rectangular areas in Fig. \ref{Fig:MSRS3} to highlight the benefits of our approach. In the red box, there are two pedestrians in the infrared image, one of whom is crossing the road. The green box has a signboard with text that is highlighted in the visible image but is entirely black in the infrared image. The red box also contains the zebra crossing from the visible image in the lower middle region. Nearly all methods emphasize the two pedestrians in the red box to varying degrees. However, two things should be mentioned. First, the body covered by  clothes in the original infrared image is not as bright as the other areas. This difference is neglected by SDDGAN and TarDAL, i.e., the whole body is equally bright, resulting in the loss of structural information. Second, many methods (e.g., FusionGAN, GANMcC, U2Fusion, SDNet) overlook the zebra crossing information from the visible image. These two issues are both avoided by our method. In addition, compared with other methods, the proposed method better preserves the information from the visible image in the green box, including brightness, color, and clarity.

\par
\begin{figure}[t]
	\centering 
	\includegraphics[width=0.49\textwidth]{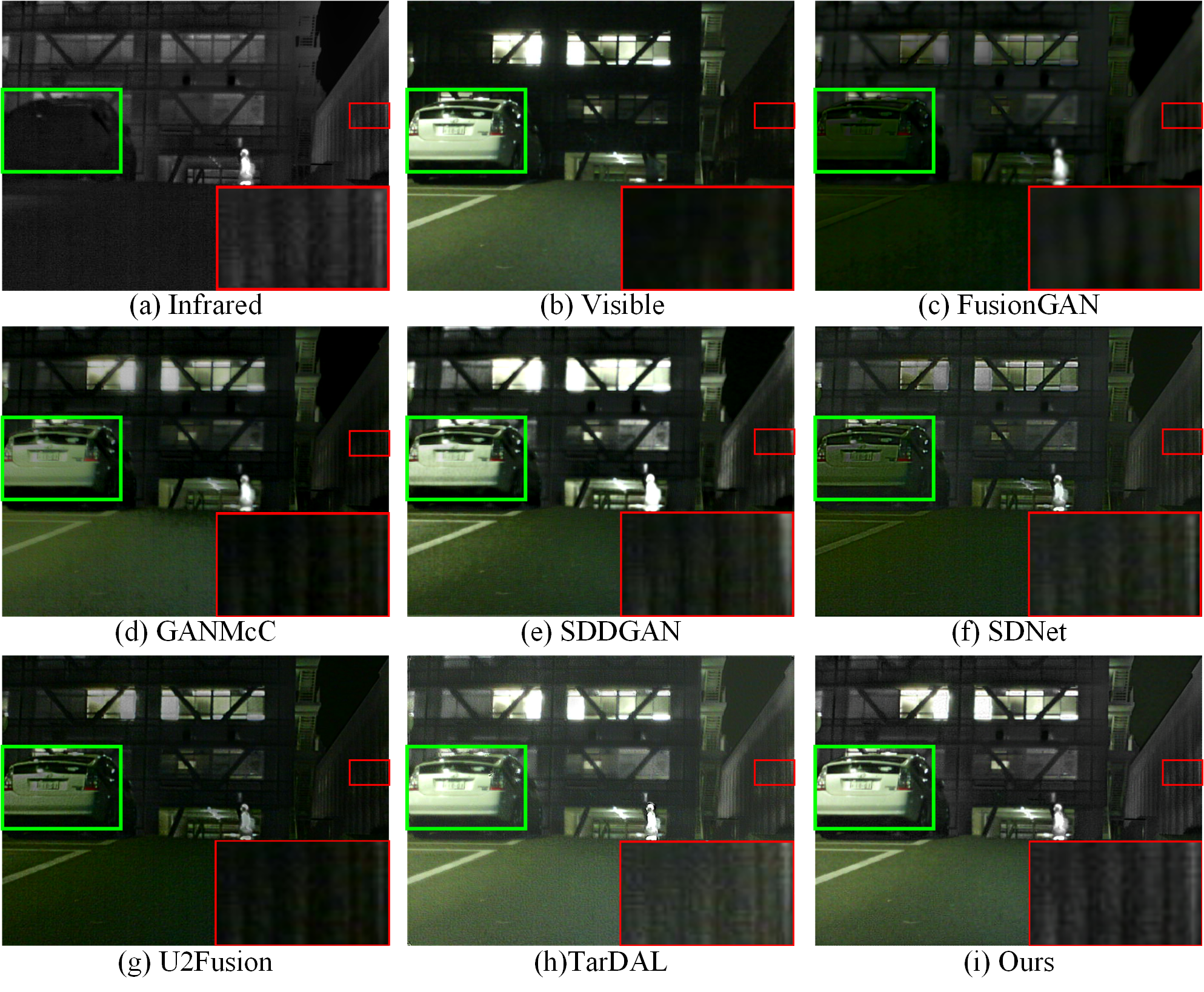} 
	\caption{Qualitative comparison of Dif-Fusion with six state-of-the-art methods on the 01061N image pair from the MSRS dataset. } 
	\label{Fig:MSRS4} 
\end{figure}

 In the second pair of nighttime images in Fig. \ref{Fig:MSRS4}, we show the fusion results in a complex lighting scene. In the infrared image, the highlighted object is a pedestrian. The area with medium brightness is located inside the windows, and the region with weak brightness is the irregular wall surfaces on the right. Areas with a lot of color and texture in the visible image, such as a white car and road surfaces, are mostly found on the left side of the image. Also, window regions in the visible image are bright. We expect the fused image to contain critical information with different levels of brightness in the infrared image. In addition, we expect to maintain the authenticity of color and texture in the visible image. To illustrate the advantages of the proposed method, we zoom in on the details of the weak signal in the red box. Going through these fused images, we observe that it is challenging to distinguish surface characteristics in the enlarged images from FusionGAN, GANMcC, U2Fusion, and TarDAL. Although the results of SDDGAN and SDNet contain this structure, they are kind of blurred or contaminated by noise. Only the fused image generated by our method is close to the original infrared image in both clarity and brightness. The images produced by FusionGAN, SDNet, GANMcC, and U2Fusion all exhibit color distortions, e.g., the white vehicle inside the green box looks green. In short, the proposed method can still preserve color fidelity in visible images and weak information from infrared images against a nighttime background by extracting complementary information from multi-channel data.

\subsubsection{Quantitative Results}
\begin{figure*}[htbp]
	\centering 
	\includegraphics[width=\textwidth]{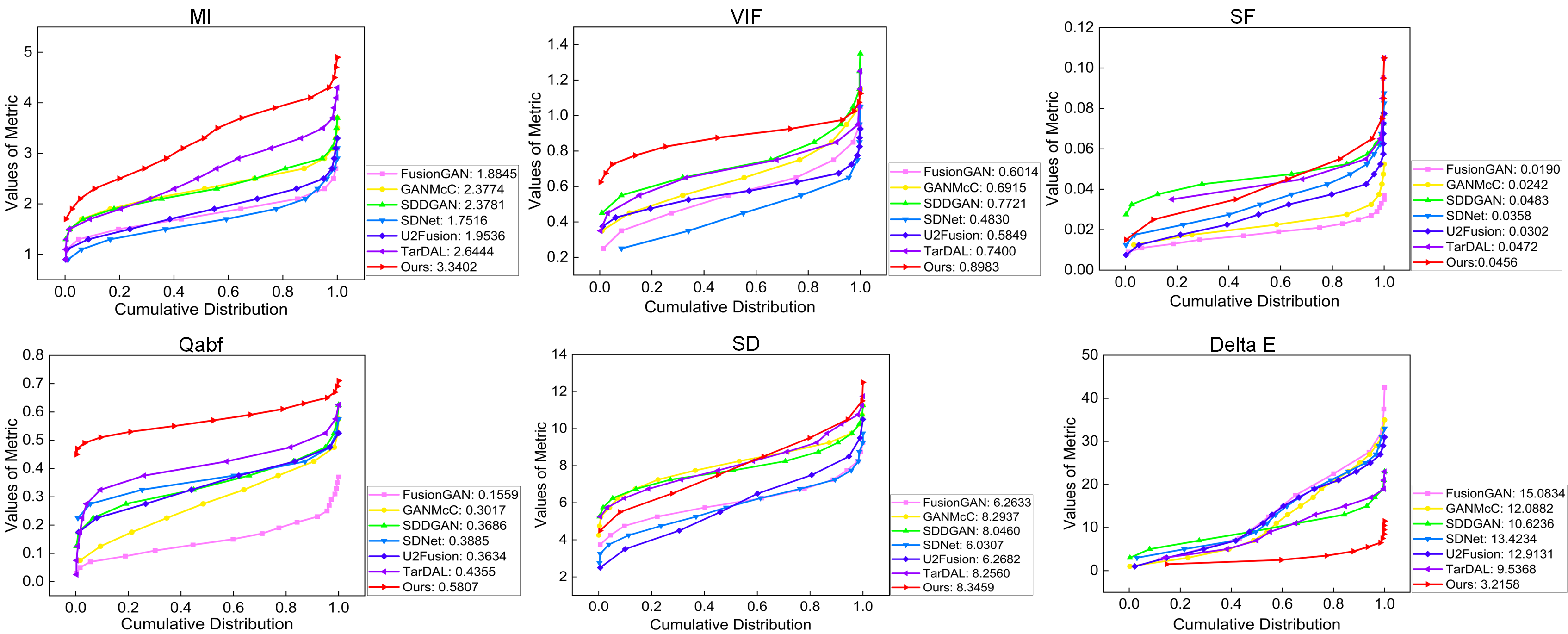} 
	\caption{Quantitative comparisons between Dif-Fusion and six state-of-the art methods on MSRS dataset with six metrics, i.e., MI, VIF, SF, Qabf, SD, and Delta E.} 
	\label{QuantitativeMSRS} 
\end{figure*}

We quantitatively compare the proposed method with six state-of-the art methods. Fig.~\ref{QuantitativeMSRS} shows the quantitative results of six statistical metrics on the MSRS dataset. We can see that our method exhibits notable benefits in five metrics (i.e., MI, VIF, Qabf, SD, and Delta E). The highest MI indicates that our method successfully transfers the most information from multi-channel source images to fused images. The best VIF shows that the fused images generated by Dif-Fusion are more in line with the human visual system. Our Dif-Fusion exhibits the best Qabf, thus more edge information is maintained. In addition, the proposed method achieves the best SD, which means that our fused images have the largest contrast. Moreover, because multi-channel complementary information is been exploited with diffusion models, our method is significantly higher than the compared method in terms of color fidelity indicator (Delta E). In the SF metric, the proposed method is just marginally inferior to SDDGAN and TarDAL.

\begin{figure}[htbp]
	\centering 
	\includegraphics[width=0.49\textwidth]{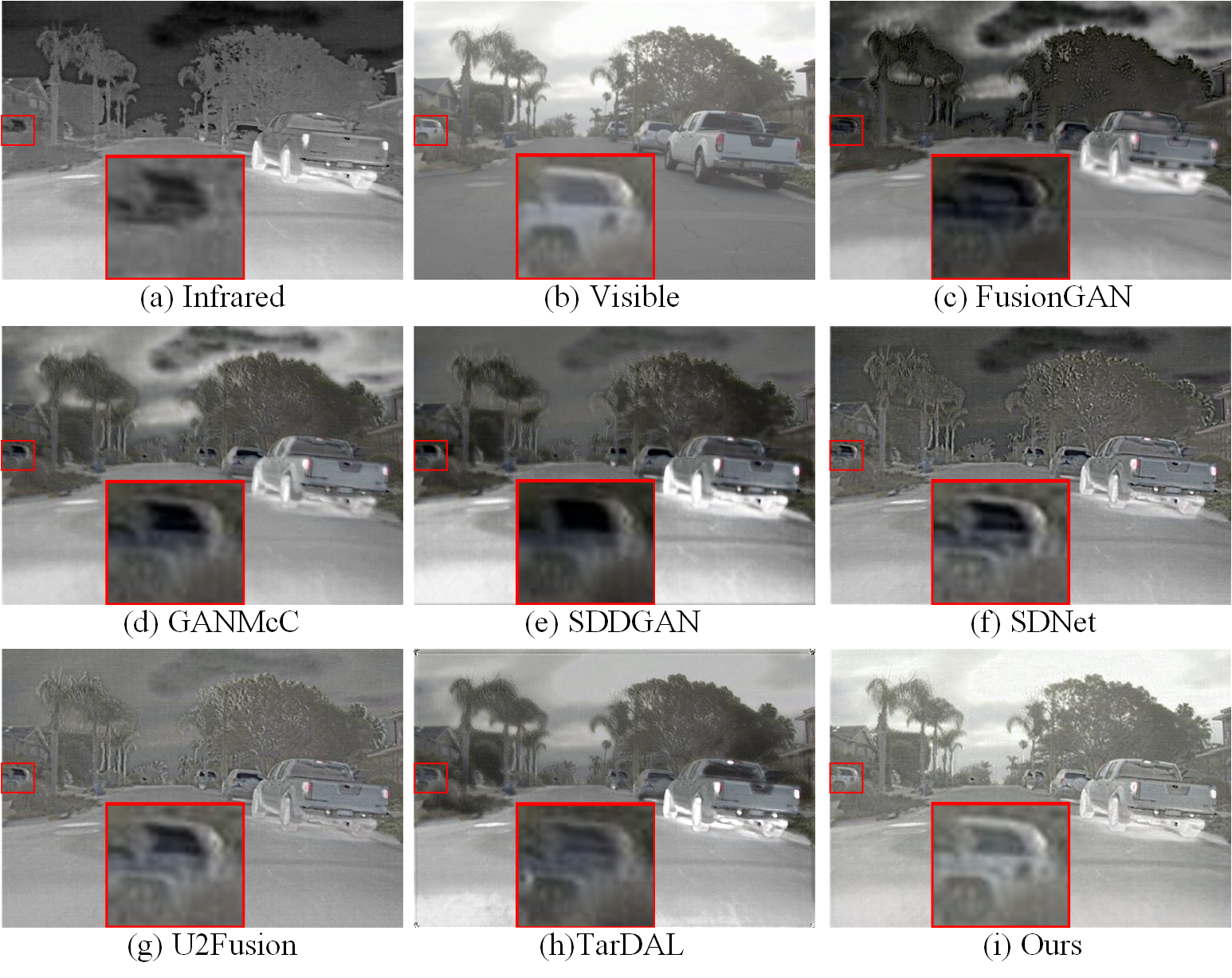} 
	\caption{Qualitative comparison of Dif-Fusion with six state-of-the-art methods on the FLIR00977 image pair from the RoadScene dataset.} 
	\label{Fig:Road1} 
\end{figure}

\subsection{Generalization Experiment}
\subsubsection{Qualitative results}
\par On the RoadScene and M3FD datasets, we test the model trained on the MSRS dataset to assess the generalization performance. The comparison methods are also tested on these two datasets. We chose one example from each dataset for an in-depth qualitative study in order to highlight the advantages of our method.
\par Fig. ~\ref{Fig:Road1} is from the RoadScene dataset. The visible image mainly consists of roads, trees, vehicles, and the sky, while the infrared image highlights the lower part of the car and part of the road surface. From the perspective of visual perception, the fused image produced by our method is the one that most resembles the original visible image. Although the bright areas in the infrared image have been preserved to some extent in the images fused using various methods, the colors of the sky and trees in the images fused by FusionGAN, GANMCC, SDDGAN, and SDNet have altered significantly. U2Fusion and TarDAL produce images with less color distortion than previous methods, but their output is blurry and lacks significant structure information (e.g., tree crown). The fused image of Dif-Fusion, in comparison, effectively preserves the salient information in the infrared image while maintaining the color and texture of prominent regions (such as the sky and trees) in the visible image. In the red rectangle, the rear of a van is enlarged. The outlines of the carriage and its wheel are muddled and cluttered in the fused image created by FusionGAN, GANMcC, SDDGAN, SDNet, U2Fusion, and TarDAL. Only our results preserve the region's color and structure details from the visible image. The ability of the proposed method to extract complementary information as well as its advantages in texture and color preservation are demonstrated by this phenomenon. 

\par
\begin{figure}[t]
	\centering 
	\includegraphics[width=0.49\textwidth]{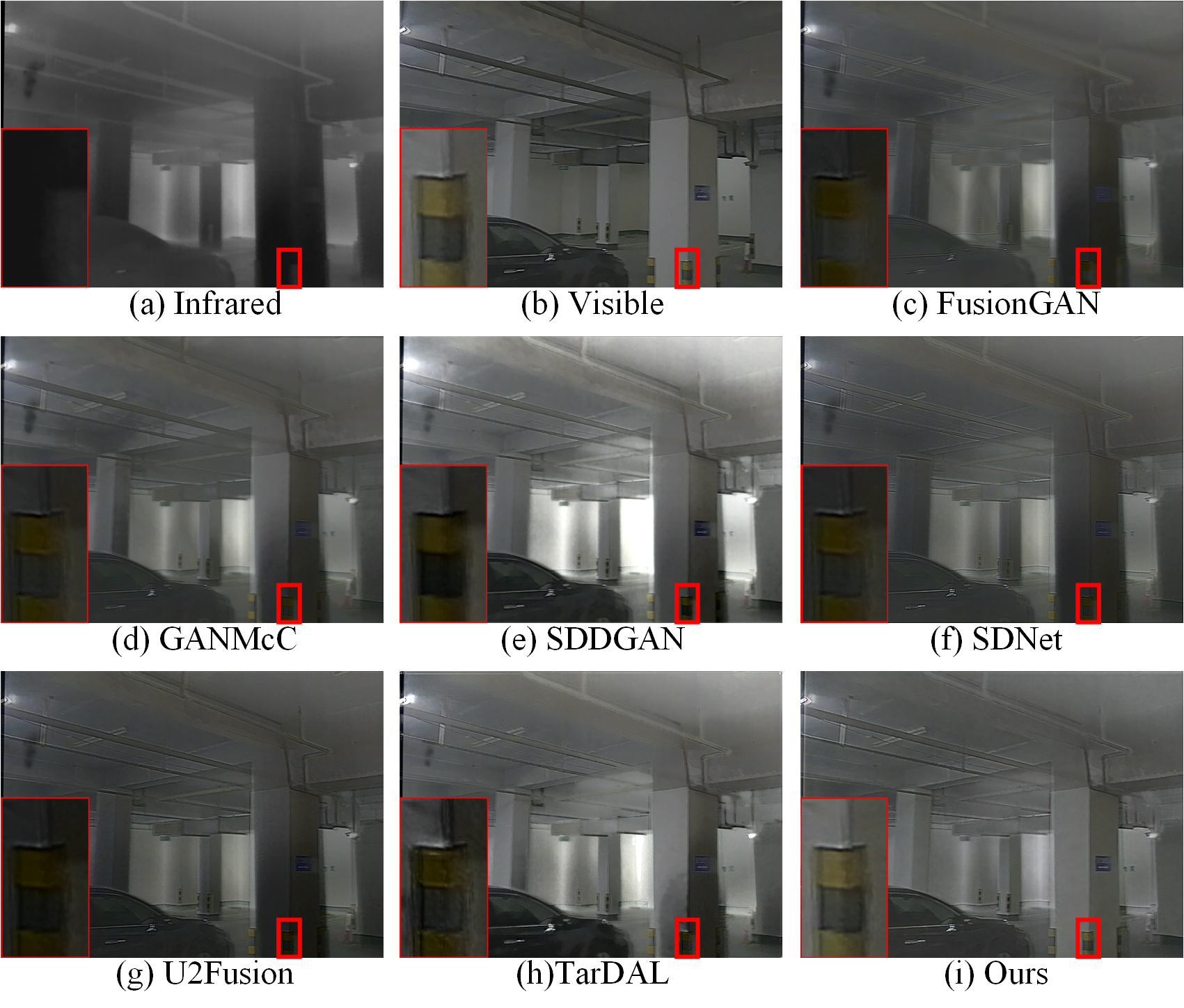} 
	\caption{Qualitative comparison of Dif-Fusion with six state-of-the-art methods on the 02757 image pair from the M3FD dataset.} 
	\label{Fig:M3F1} 
\end{figure}

 We chose an underground garage scenario from the M3FD dataset for qualitative analysis, as seen in Fig.~\ref{Fig:M3F1}. In this example, the pipeline structure and background wall are highlighted in the infrared image. First, the fused images produced by TarDAL and our approach are quite similar to the original visible image in terms of overall perception. Due to the improper combination of complementary information, FusionGAN, SDNet, and U2Fusion replace the brightness of the pillar in the visible image with the brightness of the infrared image, resulting in an excessively dark pillar on the right side of the image. GANMcC, SDDGAN, and TarDAL partially alleviate this issue. In their composite images, the pillar retains some texture and color information from the original visible image, but they are not as effective as the proposed method. Additionally, SDDGAN and TarDAL encounter the same problem as Fig.~\ref{Fig:MSRS3}, i.e., the brightness is over-enhanced, which results in the loss of wall structural information. The reflecting corner guard, which is present in the original visible image, is indicated by the red rectangle and enlarged. A thorough examination of all the enlarged views reveals that only our fused image preserves the logo's color and structural details while preserving brightness. In short, the above analysis demonstrates that the proposed method has strong generalization abilities. It can mine complementary information from multimodal data in different scenes and has good capability in texture and color retention. 

\subsubsection{Quantitative results}
We follow the previous work~\cite{tang2022image} by choosing 25 pairs of images from two datasets other than the MSRS dataset to quantitatively evaluate the generalization performance of the proposed method. Tables~\ref{M3FD25}and~\ref{RoadScene25} show the quantitative results of six statistical metrics on the M3FD and RoadScene datasets compared with six state-of-the-art methods. As shown in Table~\ref{M3FD25}, we can see that Dif-Fusion ranks first in six metrics on the M3FD dataset. The experimental results show that the fused image generated by our method has rich texture details, the highest contrast, and the best visual quality. According to Table~\ref{RoadScene25}, Dif-Fusion outperforms the compared methods in terms of VIF, Qabf, and Delta E on the RoadScene dataset. Moreover, Dif-Fusion ranks first in Delta-E on both M3FD and RoadScene datasets, which implies that the proposed method can improve color fidelity while ensuring the amount of information.

\begin{table}[t]
	\footnotesize
	\begin{center}
		\caption{Fusion quality evaluation on 25 image pairs from the M3FD dataset. Bold indicates the best results.}
		\label{M3FD25}
		\begin{tabular}{cccccccc}
			\toprule 
			Methods&MI&VIF&SF&Qabf&SD&Delta E\\
			\midrule	       
FusionGAN 	&	2.4493	&	0.5530	&	0.0413	&	0.3539	&	9.4647	&	20.0319	\\
GANMcC  	&	2.3838	&	0.6893	&	0.0365	&	0.3376	&	9.8799	&	18.1298	\\
SDDGAN	&	2.8018	&	0.8005	&	0.0477	&	0.3852	&	9.2755	&	19.3442	\\
SDNet	&	2.6705	&	0.6870	&	0.0667	&	0.5609	&	9.6364	&	20.9597	\\
U2Fusion	&	2.3374	&	0.7038	&	0.0522	&	0.5696	&	9.5848	&	16.5465	\\
TarDAL	&	2.4727	&	0.7999	&	0.0604	&	0.4428	&	9.7364	&	13.5831	\\
Ours	&	\textbf{2.9592}	&	\textbf{0.8543}	&	\textbf{0.0694}	&	\textbf{0.5771}	&	\textbf{10.0848}	&	\textbf{4.9644}	\\

			\bottomrule 
		\end{tabular}
	\end{center}
\end{table}

\begin{table}[t]
	\footnotesize
	\begin{center}
		\caption{Fusion quality evaluation on 25 image pairs from the RoadScene dataset. Bold indicates the best results.}
		\label{RoadScene25}
		\begin{tabular}{cccccccc}
			\toprule 
			Methods&MI&VIF&SF&Qabf&SD&Delta E\\
			\midrule
FusionGAN	&	2.8301	&	0.6137	&	0.0384	&	0.2985	&	10.077	&	21.2553	\\
GANMcC	&	2.8672	&	0.6957	&	0.0397	&	0.3740	&	10.1522	&	17.8119	\\
SDDGAN	&	3.0513	&	0.7121	&	0.0425	&	0.3244	&	9.5164	&	21.8514	\\
SDNet	&	3.3135	&	0.7799	&	\textbf{0.0606}	&	0.4724	&	9.8665	&	18.4423	\\
U2Fusion	&	2.7190	&	0.6710	&	0.0479	&	0.4864	&	9.7104	&	11.6658	\\
TarDAL	&	\textbf{3.3666}	&	0.8020	&	0.0570	&	0.4399	&	\textbf{10.2686}	&	12.8190	\\
Ours	&	3.3073	&	\textbf{0.8054}	&	0.0516	&	\textbf{0.5181}	&	10.1065	&	\textbf{9.1714}	\\
			\bottomrule 
		\end{tabular}
	\end{center}
\end{table}

\begin{table}[t]
	\footnotesize
	\begin{center}
		\caption{The fusion performance with and without the diffusion process on three datasets.}
		\label{ablation}
		\begin{tabular}{cccccccc}
			\toprule 
			&MI&VIF&SF&Qabf&SD&Delta E\\
			\midrule
Dataset    & \multicolumn{6}{c}{MSRS}\\
\midrule
w/o Dif.&	2.9364	&	0.8181	&	0.0376	&	0.4228	&	\textbf{8.4097}	&	4.2378	\\
w/ Dif.&	\textbf{3.3402}	&	\textbf{0.8983}	&	\textbf{0.0456}	&	\textbf{0.5807}	&	8.3459	&	\textbf{3.2158}	\\
			\midrule
Dataset    & \multicolumn{6}{c}{M3FD}\\
\midrule
w/o Dif.&	2.5870	&	0.7903	&	0.0588	&	0.4088	&	9.8860	&	9.6635	&	\\
w/ Dif.&	\textbf{2.9592}	&	\textbf{0.8543}	&	\textbf{0.0694}	&	\textbf{0.5771}	&	\textbf{10.0848}	&	\textbf{4.9644}	&	\\
			\midrule
Dataset    & \multicolumn{6}{c}{RoadScene}\\
\midrule
w/o Dif.&	3.2656	&	0.7615	&	0.0431	&	0.3478	&	8.9175	&	10.1185	\\
w/ Dif.&	\textbf{3.3073}	&	\textbf{0.8054}	&	\textbf{0.0516}	&	\textbf{0.5181}	&	\textbf{10.1065}	&	\textbf{9.1714}	\\
			\bottomrule 
		\end{tabular}
	\end{center}
\end{table}

\subsection{Ablation study}
The proposed framework adopts the diffusion models to extract multi-channel information, which improves color fidelity and visual quality with the help of multi-channel complementary information. In order to verify the effectiveness of the diffusion models, we ablate the diffusion process. More specifically, for the sake of fairness, we maintain the original network structure but removing the diffusion process. We summarize the results of the ablation study in Table~\ref{ablation}. In the MSRS dataset, after removing the diffusion process, the performance of our method decreases on five metrics (i.e., MI, VIF, SF, Qabf, and Delta E). On the M3FD dataset and the RoadScene dataset, after removing the diffusion process, the performance of our method decreases on all six metrics. It is worth noting that in the M3FD dataset, the color fidelity decreases significantly after removing the diffusion process, which indicates that the distribution of multi-channel information and the extraction of multi-channel complementary information play a very important role in color preservation.

\section{Conclusion}
In this paper, an infrared and visible image fusion method based on diffusion models is proposed to achieve multi-channel complementary information extraction and effective maintenance of color and visual quality. On the one hand, we construct the distribution of multi-channel input data in the latent space with forward and reverse diffusion process. By training a denoising network in the reverse process to predict the Gaussian noise added in the forward process, the distribution of multi-channel data is built. On the other hand, we propose a method to generate three-channel images directly. In order to preserve the gradient and intensity of the three-channel image directly, we propose multi-channel gradient and intensity losses. Moreover, in terms of fused image evaluation, in addition to the existing texture and intensity fidelity metrics, we introduce Delta E to quantify color fidelity. Extensive experiments show that Dif-Fusion is superior to existing state-of-the-art methods.

\par Overall, we investigate a framework for extracting multi-channel complementary information based on diffusion models, and try to directly generate chromatic fusion images from multi-modal input. In the future, we might explore more multi-channel information learning models and end-to-end chromatic fusion image generation methods.

\FloatBarrier

\ifCLASSOPTIONcaptionsoff
  \newpage
\fi

\bibliographystyle{IEEEtran}
\bibliography{Dif_Fusion}

\begin{IEEEbiography}[{\includegraphics[width=1in,height=1.25in,clip,keepaspectratio]{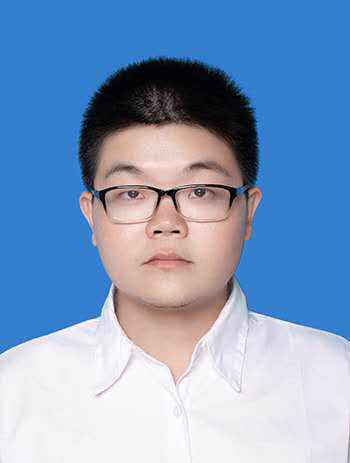}}]{Jun Yue}
	received the B.Eng. degree in geodesy from Wuhan University, Wuhan, China, in 2013 and the Ph.D. degree in GIS from Peking University, Beijing, China, in 2018. 
	
	He is currently an Assistant Professor with the School of Automation, Central South University. His research interests include satellite image understanding, pattern recognition, and few-shot learning. Dr. Yue serves as a reviewer for IEEE Transactions on Image Processing, IEEE Transactions on Neural Networks and Learning Systems, IEEE Transactions on Geoscience and Remote Sensing, ISPRS Journal of Photogrammetry and Remote Sensing, IEEE Geoscience and Remote Sensing Letters, IEEE Transactions on Biomedical Engineering, Information Fusion, Information Sciences, etc.
	
\end{IEEEbiography}

%\vspace{-20 mm}
\begin{IEEEbiography}[{\includegraphics[width=1in,height=1.25in,clip,keepaspectratio]{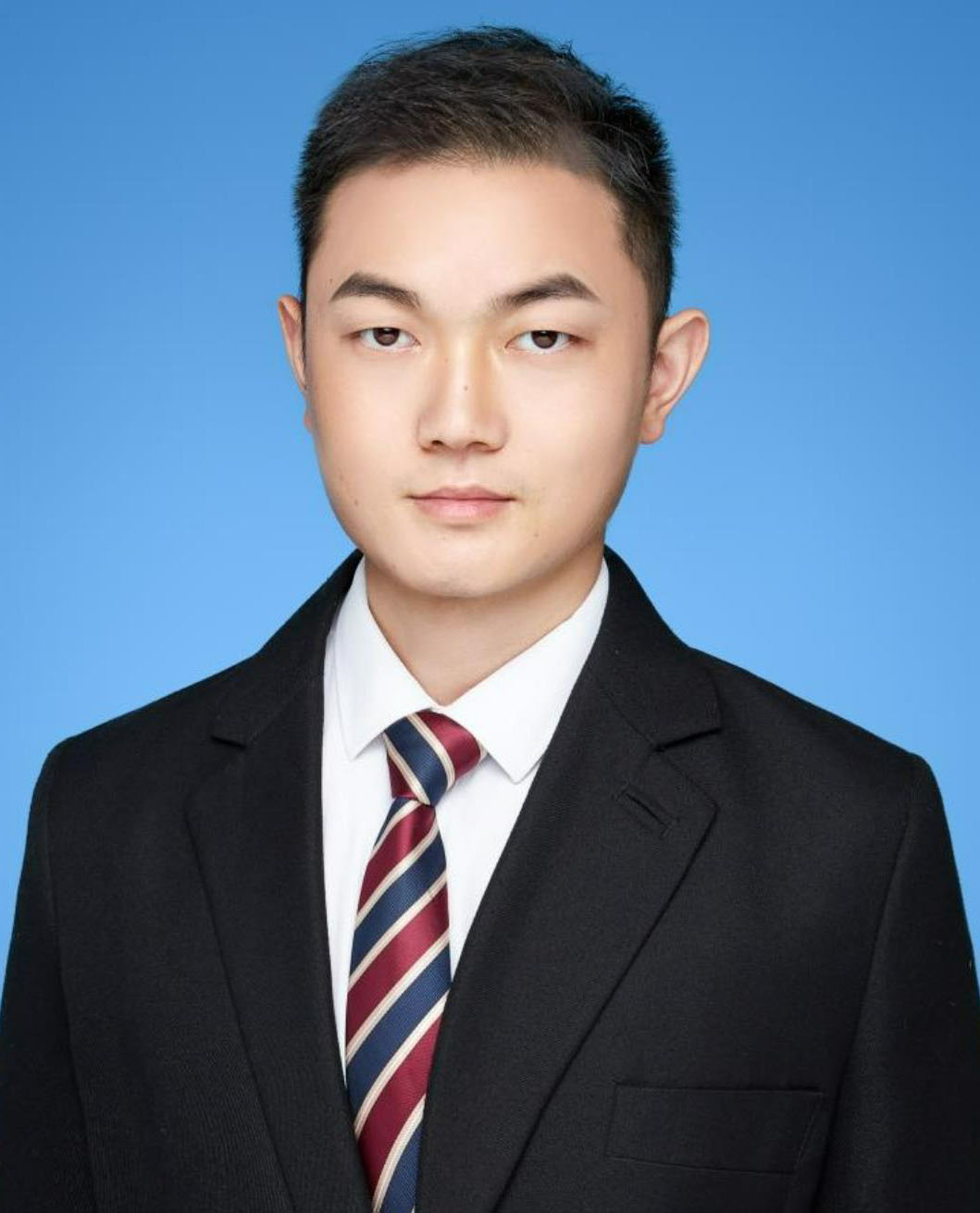}}]{Leyuan Fang}
	(Senior Member, IEEE) received the Ph.D. degree from the College of Electrical and Information Engineering, Hunan University, Changsha, China, in 2015. 
	
	From August 2016 to September 2017, he was a Postdoc Researcher with the Department of Biomedical Engineering, Duke University, Durham, NC, USA. He is currently a Professor with the College of Electrical and Information Engineering, Hunan University. His research interests include sparse representation and multi-resolution analysis in remote sensing and medical image processing. He is the associate editors of IEEE Transactions on Image Processing, IEEE Transactions on Geoscience and Remote Sensing, IEEE Transactions on Neural Networks and Learning Systems, and Neurocomputing. He was a recipient of one 2nd-Grade National Award at the Nature and Science Progress of China in 2019.  
\end{IEEEbiography}
%\vspace{-20 mm}

\begin{IEEEbiography}[{\includegraphics[width=1in,height=1.25in,clip,keepaspectratio]{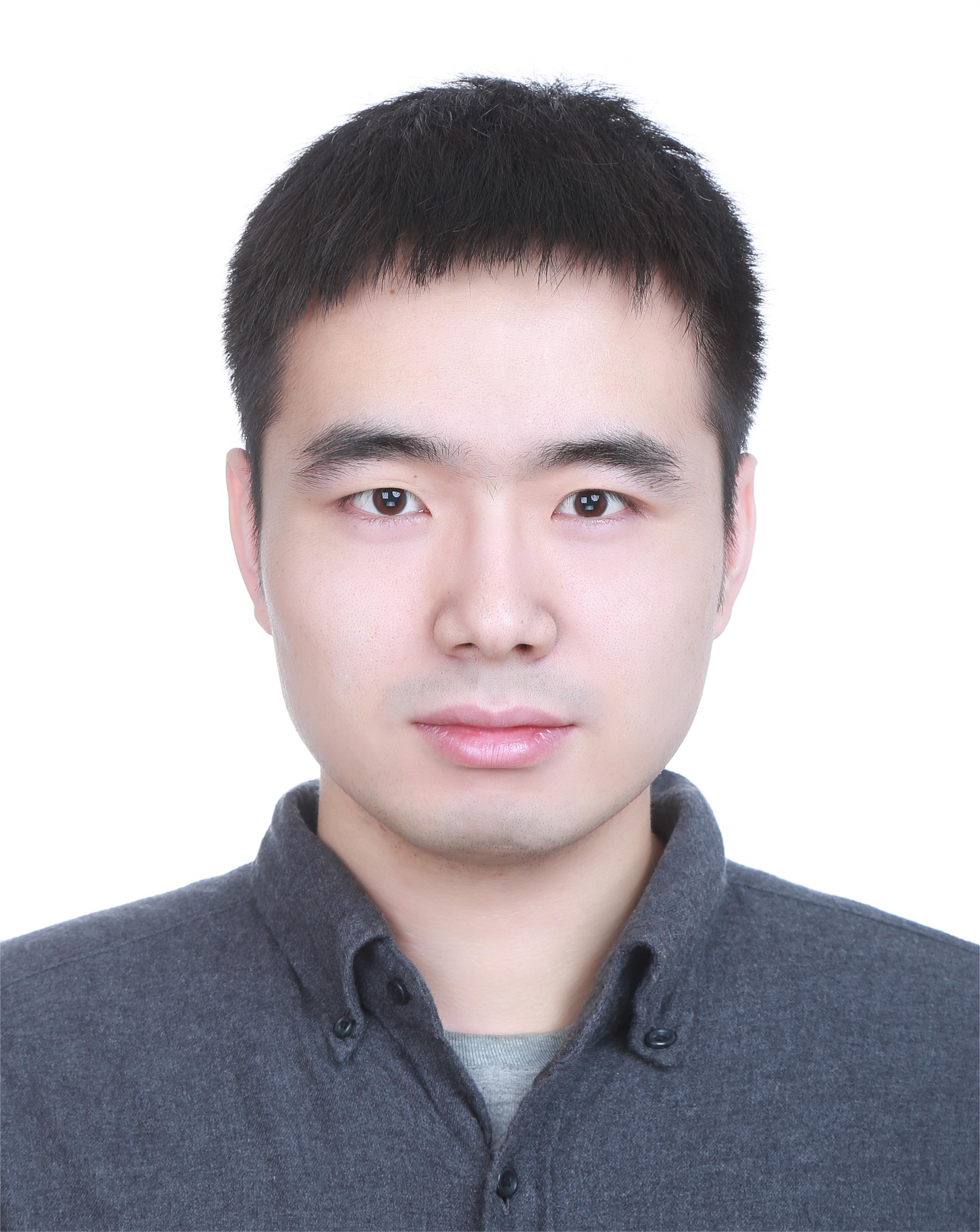}}]{Shaobo Xia} received the bachelor's degree in geodesy and geomatics from the School of Geodesy and Geomatics, Wuhan University, Wuhan, China, in 2013, the master's degree in cartography and geographic information systems from the Institute of Remote Sensing and Digital Earth, Chinese Academy of Sciences, Beijing, China, in 2016, and the Ph.D. degree in geomatics from the University of Calgary, Calgary, AB, Canada, in 2020.
	
	He is an Assistant Professor with with the Department of Geomatics Engineering, Changsha University of Science and Technology, Changsha, China. His research interests include point cloud processing and remote sensing.
\end{IEEEbiography}

\begin{IEEEbiography}[{\includegraphics[width=1in,height=1.25in,clip,keepaspectratio]{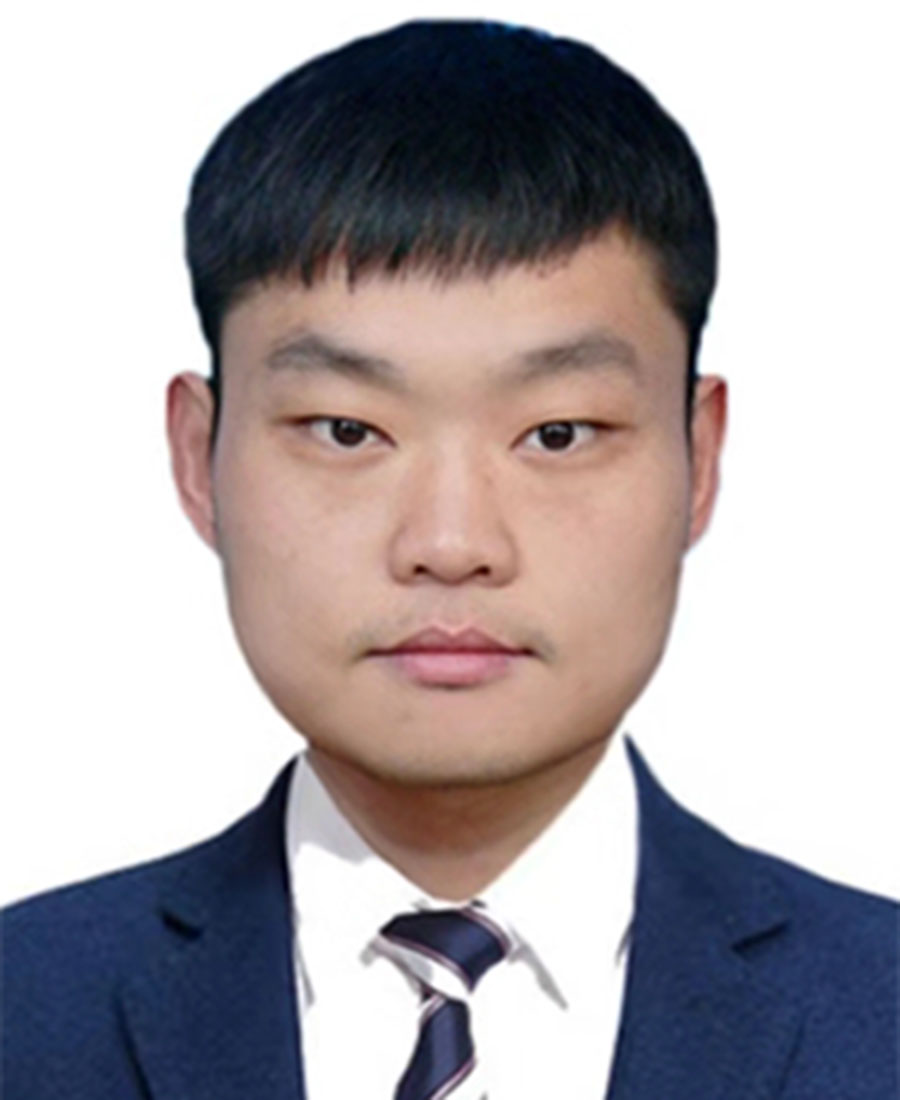}}]{Yue Deng} (Senior Member, IEEE) received the Ph.D. degree (Hons.) in control science and engineering from the Department of Automation, Tsinghua University, Beijing, China, in 2013.

He is currently a Professor with the School of Astronautics, Beihang University, Beijing, China. His research interests include machine learning, signal processing, and computational biology.
\end{IEEEbiography}

\begin{IEEEbiography}[{\includegraphics[width=1in,height=1.25in,clip,keepaspectratio]{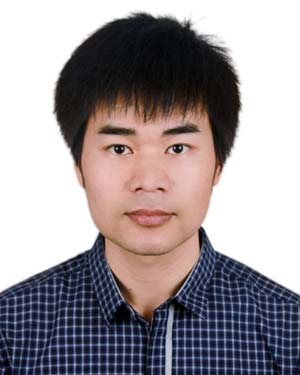}}]{Jiayi Ma} (Senior Member, IEEE) received the B.S. degree in information and computing science and the Ph.D. degree in control science and engineering from the Huazhong University of Science and Technology, Wuhan, China, in 2008 and 2014, respectively. 

He is currently a Professor with the Electronic Information School, Wuhan University, Wuhan. He has authored or coauthored more than 300 refereed journal and conference papers, including IEEE TPAMI/TIP, IJCV, CVPR, ICCV, ECCV. His research interests include computer vision, machine learning, and robotics. Dr. Ma has been identified during 2019–2022 Highly Cited Researcher lists from the Web of Science Group. He is an Area Editor of Information Fusion and an Associate Editor of Neurocomputing.
\end{IEEEbiography}

\FloatBarrier

\end{document}